\pdfoutput=1

\documentclass[11pt]{article}

\usepackage[]{EACL2023}

\usepackage{times}
\usepackage{latexsym}
\usepackage{hyperref}
\usepackage{graphicx}
\usepackage{float}
\graphicspath{ {./images/} }
\usepackage{soul}

\usepackage[T1]{fontenc}

\usepackage[utf8]{inputenc}

\usepackage{microtype}

\usepackage{inconsolata}

\title{
Croatian Film Review Dataset (Cro-FiReDa): A Sentiment Annotated Dataset of Film Reviews
}

\author{ Gaurish Thakkar \and  Nives Mikelić Preradović  \and  Marko Tadić \\
  Faculty of Humanities and Social Sciences, University of Zagreb, Zagreb 10000, Croatia\\ \texttt{gthakkar@m.ffzg.hr, nmikelic@ffzg.hr, marko.tadic@ffzg.hr } }

\begin{document}
\maketitle
\begin{abstract}
This paper introduces Cro-FiReDa, a sentiment-annotated dataset for Croatian in the domain of movie reviews. The dataset, which contains over 10,000 sentences, has been annotated at the sentence level. In addition to presenting the overall annotation process, we also present benchmark results based on the transformer-based fine-tuning approach. 
\end{abstract}

\section{Introduction}
The goal of sentiment analysis is to classify the polarity of text (\emph{e.g.}, positive, negative, neutral, or mixed). In this paper, we describe the process of annotating a sentiment analysis dataset in Croatian. As shown in the example below, the label indicates the sentiment polarity of the text.
\begin{itemize}
    \item[\textbf{Hr}] “I bio sam zadivljen i tijekom finalne borbene scene .”
    \item[\textbf{En}] “And I was also amazed during the final battle scene.”
    \item \textbf{Label} : positive
\end{itemize}
Croatian is a low-resource language in terms of sentiment analysis resources. There is currently no Croatian dataset for the domain of movie reviews. The dataset presented here is the first sentiment movie review dataset. 
The texts for the annotation campaign are taken from the Croatian movie review website and cover multiple genres, namely adventure, series (serija), and sci-fi. In addition to the other metadata described below, the website includes a summary of the entire text of the author's review. The dataset, annotation guidelines, trained models, and associated code will be made available to the public. In this work, we describe our entire workflow for creating the resource. We also present the experimental scores for the sentiment analysis task using pre-trained transformer models.

The rest of the paper is structured as follows: In Section 2, we review the related work on the dataset with regard to its annotation as well as modelling. In Section 3, we describe the annotation process in detail. In Section 4, we present the statistics of the annotated dataset before presenting the baseline scores in Section 5. We complete the paper with the conclusion, discussion, and future work in Section 6.

\section{Related Work}
In this section, we will highlight the related work on resources and models for sentiment analysis.
Sentiment analysis is a well-researched field, and there are a number of resources for various languages, such as English \citep{maas-EtAl:2011:ACL-HLT2011,Pang+Lee:05a,marc_reviews}, German \citep{cieliebak-etal-2017-twitter,sanger-etal-2016-scare, clematide-etal-2012-mlsa}, French \citep{apidianaki:hal-01838536}, and Italian \citep{basile-nissim-2013-sentiment}. There are few resources available for Croatian sentiment analysis. The stance (and sentiment) annotated dataset \citep{bosnjak-karan-2019-data} contains comments submitted by users for online news articles. \citet{app10175993} created a dataset for sentiment analysis of Croatian news articles and performed zero-shot classification using Slovene resources. \citet{zhou-etal-2015-wikipedia} performed multiple levels of sentiment analysis on multilingual Wikipedia articles using machine translation. \citet{ohman-etal-2020-xed} compiled a parallel dataset for sentiment and emotion analysis based on movie subtitles. The dataset was created by manually annotating 25K Finnish and 30K English sentences, which were then projected onto 30 other languages, including Croatian. \citet{agic-etal-2010-towards} presented rule-based annotated Croatian news articles in the finance domain that captured the general sentiment of the text. \citet{rotim-snajder-2017-comparison} compiled a dataset of gaming review text spans in Croatian that were tagged with positive and negative labels. There are also a few Croatian sentiment lexicons, such as those developed by \citet{ljubesic-etal-2020-lilah, glavas-etal-2012-experiments}. The ParlaSent-BCS \citep{11356/1585} dataset is another resource that has Croatian sentences in parliamentary debates tagged with sentiment polarity. \citet{tikhonov-etal-2022-eenlp} provide an overview of existing resources for East European languages, including Croatian.

\section{Text and Annotations}
In this section, we describe our annotation procedure in greater depth. First, we describe the backgrounds of the annotators. Second, the guidelines and methodology for annotation are explained. Third, the statistical aspects of the dataset are discussed. 

\subsection{Annotation Procedure}
\begin{figure}[h]
\centering
\includegraphics[width=0.40\textwidth]{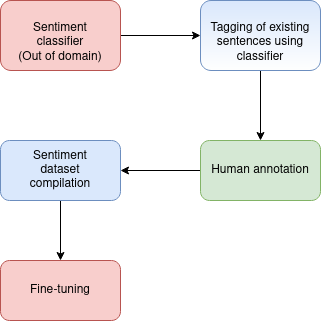}
\caption{The dataset creation process.}
    \label{fig:fig1}
\end{figure}

The task is defined as a sentence-level sentiment task in which each sentence in the training set is annotated with a single label. The dataset consists of professional reviews from the Croatian movie review website\footnote{https://www.recenzijefilmova.com/}. The adventure, TV series, and science fiction (sci-fi) genres were chosen as sub-categories. Each review instance is accompanied by the following data fields:
\begin{enumerate}
    \item \textbf{Review}: the text written by the professional reviewer.
    \item \textbf{First impression}: short summary of the overall review.
    \item \textbf{Overall assessment}: the score assigned by the reviewer. The reviewers rate the film on different scales, the scores range from (0-10) to (1-5) stars.
    \item \textbf{Date}: date of the review.
\end{enumerate}

In addition, the review text has formatted information about the title, IMDB rating, producers, actors, directors, genres, and date of release. The dataset contains a total of 216 adventure-related reviews, 114 sci-fi reviews, and 76 series reviews. We framed the sentiment annotation task as a sentence-level label correction task. The overall methodology is presented in Figure \ref{fig:fig1}. Each review has undergone sentence segmentation, in which the entire review has been broken down into individual sentences. All reviews were sorted by sentence length and divided into groups so that each annotator received an equal amount of sentences, but at the same time, no annotator received partial review text. This was done to make sure that no student received a partial review. An empirical method was used to determine the N=23 groups. A minimum of three (and a maximum of five) annotators have annotated a single sentence. Each review was pre-annotated using the deep-learning sentiment classification model \citep{thakkar2021multi}. The classification model was trained using the SentiNews dataset, which is composed of Croatian and Slovenian news articles in a multitask setup, and has reported an F1-score of 63.86. This step has sped up the annotation process, as annotators are no longer required to tag the sentence from scratch, but only correct the tag if it is incorrect. A total of 82 students participated in the study. All the annotators were undergraduate students of linguistics and informatics between the ages of 22 and 24. All the annotators were native Croatian speakers. The final label for a sentence is chosen by a majority vote.

\subsection{Annotation Scheme}
The guidelines for the annotation were largely adopted from \citet{mohammad-2016-practical}. Learners were presented with five categories of sentiment: 1—negative, 2—neutral, 3—positive, 4—mixed, and 5—other/sarcasm. Evidently, the negative review is labelled as negative, while the positive review is labelled “positive”. The release date and genre of the film are categorised as neutral facts. Sentences that have both positive and negative connotations are classified as “mixed”. If figurative language exists, it is labelled as “other/sarcasm”. The annotation guidelines describe each instance of the label with multiple examples.

\subsection{Web Interface}
All of our annotation tasks used the online tool INCEpTION \cite{tubiblio106270} because it enables simple semantic annotations. The platform simplifies the administration of annotation projects involving many annotators. Because we did not want participants to see each other's work, each group of students was assigned a separate project. Each user was subsequently able to view only the files assigned to him or her after logging into the system with his or her credentials. Before moving on to the next document, each user would perform the annotation process and lock the document. The locking mechanism signified the document's completion and allowed us to monitor its completion status and overall work status. Each student has averaged four hours on the assignment.

\subsection{Inter-annotator agreement}
Using Fleiss Kappa, we have measured the inter-annotator agreement of the dataset across multiple groups. The scores suggest moderate (0.41-0.60) to substantial (0.61-0.80) levels of annotator agreement. Table \ref{tab:table1} lists the agreement for every label. During the phase of judging, the annotators were required to report any uncertainties. The majority of queries pertained to metadata present in the review text, such as the title. There were 843 disagreements in which there was no clear majority winner. These sentences were characterised by conditionals or mixed sentiments and were filtered out as they were not additionally annotated by anyone and will be taken up for future work.

\begin{itemize}
    \item[\textbf{Hr}] “Za one koji vole ovu vrstu filma , trebali biste biti u mogućnosti uživati , ali za lojalnog ljubitelja izvornog filma , ovaj se može vidjeti kao još jedan od najljepših ili najmanje omiljenih .”	
    \item[\textbf{En}] “For those who like this type of film, you should be able to enjoy it, but for a loyal fan of the original film, this one can be seen as another of the best or least favorite.”
\end{itemize}

\subsection{Corpus Statistics}
Table \ref{tab:table1} shows the statistics for the final sentiment annotated dataset. Out of 10,464 sentences, we have 59 percent neutral statements. This is clear because the majority of the text contains factual information about the movie/series. There are a total of 875 reviews that have text summaries associated with the main text. The mean number of space-separated tokens for review text and summary is 731 and 47, respectively.

\begin{table}[h]
\centering
\begin{tabular}{lcc}
\hline
\textbf{Label} & \textbf{\# of instances}& \textbf{agreement} \\
\hline
neutral & 6205 & 0.51\\
positive & 2031 & 0.53 \\
negative & 1290 & 0.42 \\\hline
mixed & 862 & 0.30 \\
sarcasm & 76 & 0.04 \\\hline\hline
total &10464\\\hline\hline
\end{tabular}
\caption{Statistics of the sentiment dataset. Numbers represent sentences. Kappa statistics for each label }
\label{tab:table1}
\end{table}

\subsection{Dataset Analysis}
Out of 10,388 samples, around 2,257 instances retained their original classification tag. The remaining 8,131, which constitute around 78 percent of the final dataset, were modified by the annotators. In these modifications, more than 50 percent of the changes (4,813 instances) were from negative label to neutral, followed by a positive to neutral annotation change (1,053 instances). The sentences that changed from non-neutral to neutral were mostly informative, similar to title sentences with polar words. We also sampled a few random reviews and checked the polarity of the individual sentences in the review, ignoring the neutral sentences. This number of positive and negative sentences does hint at the possibility of a relationship with the overall rating of the review given by the reviewer. For instance, if there were an equal number of positive and negative sentences, the movie would receive a 3/5 or 5/10 rating. On the other side, if the review contains more compliments, it will receive a rating higher than 3. Exactly 654 sentences in the groups received the same annotated class provided by the authors.

\section{Experiments}

\subsection{Experimental Setup}
We performed experiments for the task of sentiment analysis. To benchmark the dataset on sentiment classification, we use the fine-tuning approach proposed by \citet{devlin-etal-2019-bert}. We used the CroSloEngual BERT \citep{ulcar-robnik2020finest} as our contextualised pre-trained language model and performed fine-tuning using a softmax classification head. CroSloEngual
BERT was trained on corpora from Croatian, Slovenian, and English languages with a total of 5.9 billion tokens. For training, only the positive, negative, and neutral class instances were used. We divided the dataset into train tests in an 80:20 ratio and used 10\% of the train set for development. We used a learning rate of 1e-05 and weight decay of 0.02 with early stopping on evaluation loss with patience of 4. A batch size of 16 was used during training. In addition, a hidden dropout and attention of 0.2 were used as regularization constants. Each of the experiments used a GPU with 24 GB of VRAM. Each epoch of sentiment training lasted longer than 20 minutes.
In addition, we also present the results utilising the three strategies described in \citet{app10175993}. The reported approaches employ 10-fold cross-validation for training stage. A hidden layer (768,250), ReLU activation, and a softmax classification layer are used in the second and third methods (250, number of classes).
The overlapped long texts used in the second technique are used to build an oversampled dataset.
The third method averages all the vectors corresponding to the overlapping sentences, rather than oversampling them. The vectors are subsequently sent through a ReLU-equipped two-layered classification head.  

\section{Results}
The scores for the sentiment task are reported using the F1, and accuracy (macro) metrics. In the case of fine-tuning setup, each experiment was performed five times with different random seeds, and the mean of all the scores is reported in Table \ref{tab:table3}. We also tested the XLM-RoBERTa \citep{DBLP:journals/corr/abs-1911-02116} and classla/bcms-bertic \citep{ljubesic-lauc-2021-bertic} language models, both of which were pre-trained in Croatian, but the results were no better than the CroSloEngual BERT. All the scores are comparable, as the final scores are reported on the same test set that was held out during the training phase.

\begin{table*}[h]
\centering
\begin{tabular}{ccc}
\hline
\textbf{Configuration} & \textbf{F1} & \textbf{Accuracy}\\
\hline
FT $\dagger$ &  79.78 (0.008) & 84.71 (0.006) \\

\hline
CV $\diamond$   &  71.19 ( 0.007) & 80.43 (0.003)   \\

\hline
Sampling average  $\diamond$  &  70.84 (0.003) & 80.13 (0.002) \\

\hline

CV sampling  $\diamond$   & 70.69 (0.005) & 80.18 (0.002)   \\
\hline

\end{tabular}
\caption{Results of the experiments. $\dagger$: \citet{devlin-etal-2019-bert}. $\diamond$: Methods reported in \citet{app10175993}}
\label{tab:table3}
\end{table*}

\subsection{Error Analysis}
A manual error analysis points to two major categories of errors. First, there are instances in the annotated set that have polar labels for metadata about the movie. Second, the trained model also has problems dealing with conditionals. Two instances are provided below.
\begin{enumerate}
    \item 
        \begin{itemize}
            \item[\textbf{Hr}] “Ako tražite nešto zbog čega razmišljate , a usredotočujete se dosta na odnos , onda bi vas ova serija trebala zabaviti.”
            \item[\textbf{En}] “If you are looking for something to make you think and focus a lot on the relationship, then this series should entertain you.”
        \end{itemize} 
    \item  
        \begin{itemize}
            \item[\textbf{Hr}] “Kad bi samo satovi znanosti u školi bili zabavni.”
            \item[\textbf{En}] “If only science lessons at school were fun.”
        \end{itemize} 
\end{enumerate}

\subsection{Discussion}
All the annotators were presented with a questionnaire to be answered upon the completion of the task. The questionnaire included basic questions like how much time was required on average, good and bad experiences, as well as suggestions for future enhancements.
Apart from the enhancement of the user interface for the annotation tool, one common request was to include neutral-positive and neutral-negative. These were mainly sentences that were objective in nature, but invoked sentiment. For example,
\begin{itemize}
    \item[\textbf{Hr}]  “Ocjena na IMDb.com mu je 6,4 / 10 , a na Tomatoesima malih 36\%.” 
    \item[\textbf{En}]  “The rating on IMDb.com is 6.4 out of 10, and on Tomatoes it is 36\%.”
\end{itemize}
This was one of the sentences in which two annotators had tagged it neutral, while the other two had tagged it with a negative label.

\section{Conclusion and Future Work}
With this paper, we have presented the sentiment annotated movie review dataset for Croatian. We performed experiments using curated datasets for the sentiment analysis task for the Croatian language. Out of 21 unique categories of film reviews, to name a few, we have processed only three categories (adventure, series (serija), and sci-fi). In the future, we would like to use the gold-annotated dataset in a distant-supervised learning regime to perform sentiment classification on all the non-annotated reviews. Another area of research would be to formally evaluate how pre-suggestions of the model before manual annotation could influence annotators' decisions. For example, a systematic comparison of labelling sentences from scratch versus allowing people to correct/retain automated labels could be conducted. In addition, we would like to experiment with mixed and sarcasm-tagged sentences. The dataset also contains metadata, such as genres and document-level sentiment ratings, which can be explored in the future. 

\section*{Acknowledgement}
The work presented in this paper has received funding from the European Union’s Horizon 2020 research and innovation program under the Marie Skłodowska-Curie grant agreement no. 812997 and under the name CLEOPATRA (Cross-lingual Event-centric Open Analytics Research Academy)

\section*{Limitations}
Although the current dataset mainly covers the genres of sci-fi, adventure, and series, there are other genres (games and books) that are missing from the dataset. The models were trained on a 24 GB GPU. Hence, we expect this could limit reproducibility. The downside of the presented approach is the decision to use an existing classifier to pre-annotate the texts. The suggestions could bias the students.


\section*{Ethics Statement}
The annotated dataset reported in this paper involved manual effort. This is an output of the annotation campaign conducted with students of linguistics and informatics in order to aid in learning about sentiment analysis as part of their coursework. The students were compensated with course credits at the end of the campaign.

\bibliography{custom}
\bibliographystyle{acl_natbib}

\appendix

\section{Appendix}
\label{sec:appendix}

\begin{table}[h]
\centering
\begin{tabular}{llc}
\hline
\textbf{Task} & \textbf{Metric} & \textbf{Value}\\
\hline
sentiment & learning rate & 1e-5  \\
{} & weight decay & 0.02 \\ 
{} & batch size &  16 \\
{} & epochs & 10   \\
\hline

\\ \hline
\end{tabular}
\caption{List of hyperparameters, model parameters and their values used during the experiments.}
\label{tab:table4}
\end{table}

\begin{table}[!ht]
\centering
\begin{tabular}{cc}
\hline
\textbf{\#} & \textbf{Category} \\
\hline
1 & adventure  \\
2 & new-films \\
3 & biography  \\
4 & comedy \\
5 & documentary \\
6 & sci-fi \\
7 & thriller \\
8 & sport \\
9 & war \\
10 & western \\
11 & mystery \\
12 & crime \\
13 & family \\
14 & drama \\
15 & music \\
16 & history \\
17 & action \\
18 & romance \\
19 & animation \\
20 & fantasy \\
21 & horror   \\
22 & series  \\\hline
\end{tabular}
\caption{List of categories.}
\label{tab:table5}
\end{table}

\end{document}